\documentclass[letterpaper, 10 pt, conference]{ieeeconf}  % Comment this line out if you need a4paper

\usepackage[noadjust]{cite}
\usepackage{amsfonts}
\usepackage{amsmath}
\usepackage{xcolor}
\usepackage{amssymb} % for \smallsetminus
\usepackage{diagbox} % for \backslashbox
\usepackage{lipsum}  % for random text, to be removed
\usepackage{multirow} % for \multirow
\usepackage{url}
\usepackage{graphicx}

\usepackage{algorithmic}
\usepackage{algorithm}

\IEEEoverridecommandlockouts                              % This command is only needed if 
\algsetup{linenosize=\tiny}
\title{\LARGE \bf
B$^2$F-Map: Crowd-sourced Mapping with Bayesian B-spline Fusion}
 \author{Yiping Xie$^{1,2*}$, Yuxuan Xia$^{3*}$, Erik Stenborg$^{1}$, Junsheng Fu$^{1}$, Axel Beauvisage$^{1}$,\\ Gabriel E. Garcia$^{1}$, Tianyu Wu$^{1,4}$ and Gustaf Hendeby$^{2}$% <-this % stops a space
 \thanks{*This work was supported in part by the Wallenberg AI, Autonomous Systems and Software Program (WASP) funded by the Knut and Alice Wallenberg Foundation. (Corresponding author: Yiping Xie.)}% <-this % stops a space
 \thanks{$^{1}$Zenseact, Lindholmspiren 2, Gothenburg, Sweden.
         {\tt\small \{yiping.xie, erik.stenborg, junsheng.fu, axel.beauvisage, gabriel.garcia-jaime, tianyu.wu\}@zenseact.com}}%
 \thanks{$^{2}$ Department of Electrical Engineering,
Linköping University, Linköping, Sweden.
         {\tt\small \{yiping.xie, gustaf.hendeby\}@liu.se}}%
\thanks{$^{3}$Yuxuan Xia was a Post-doc with Zenseact, and he is now with the
Department of Automation and Perception, Shanghai Jiao Tong University,
Shanghai, China.
         {\tt\small yuxuan.xia@sjtu.edu.cn}}%
    \thanks{$^{4}$Department of Electrical Engineering, Chalmers University of Technology, Gothenburg, Sweden.
         {\tt\small wuti@chalmers.se}}%     
         \thanks{$^{*}$Equal contributions.}
 }

\begin{document}

\maketitle
\thispagestyle{empty}
\pagestyle{empty}

%%%%%%%%%%%%%%%%%%%%%%%%%%%%%%%%%%%%%%%%%%%%%%%%%%%%%%%%%%%%%%%%%%%%%%%%%%%%%%%%
\begin{abstract}
Crowd-sourced mapping offers a scalable alternative to creating maps using traditional survey vehicles. Yet, existing methods either rely on prior high-definition (HD) maps or neglect uncertainties in the map fusion. In this work, we present a complete pipeline for HD map generation using production vehicles equipped only with a monocular camera, consumer-grade GNSS, and IMU. Our approach includes on-cloud localization using lightweight standard-definition maps, on-vehicle mapping via an extended object trajectory (EOT) Poisson multi-Bernoulli (PMB) filter with Gibbs sampling, and on-cloud multi-drive optimization and Bayesian map fusion. We represent the lane lines using B-splines, where each B-spline is parameterized by a sequence of Gaussian distributed control points, and propose a novel Bayesian fusion framework for B-spline trajectories with differing density representation, enabling principled handling of uncertainties. We evaluate our proposed approach, B$^2$F-Map, on large-scale real-world datasets collected across diverse driving conditions and demonstrate that our method is able to produce geometrically consistent lane-level maps.
\end{abstract}\vspace{-4pt}

%%%%%%%%%%%%%%%%%%%%%%%%%%%%%%%%%%%%%%%%%%%%%%%%%%%%%%%%%%%%%%%%%%%%%%%%%%%%%%%%
\section{INTRODUCTION}
High-definition (HD) maps are fundamental for enabling safe and reliable autonomous driving. Traditionally, these maps are generated by expensive survey vehicles equipped with high-precision sensors such as high-grade inertial sensors, RTK-GPS, and LiDAR, followed by labor-intensive annotations. In addition to high operational costs, traditional approaches also struggle with scalability and timely updates. Recently, crowd-sourcing, which leverages data from widely distributed production vehicles equipped with low-cost and consumer-grade sensors, has emerged as a promising alternative to overcome these limitations, offering a path towards more cost-effective, scalable, and frequently updated maps. 

While recent efforts have explored pure vision-based crowd-sourced mapping, many still rely on certain prior assumptions or address specific sub-problems, instead of a full, automatic crowd-sourcing pipeline. For example, \cite{bellusci2024semantic} and \cite{Yuxuan2024Fusion} only demonstrated the map generation pipeline from single-drive data without the module to align and fuse maps from multiple drives. MapCVV \cite{Pengxin2024RAL} requires a base map, produced by vehicles equipped with RTK-GPS \cite{Tong2021ICRA}, for its on-vehicle localization module. Similarly, \cite{Jian2023ITS} requires an existing HD map for the initial mapping match. 

Another research gap is map fusion, specifically the aggregation of vectorized lane lines from different drives. Many works \cite{Onkar2017IROS,Jian2023ITS,Zebang2025ITSC} completely ignored the uncertainties of the vectorized lane lines, including both mapping and localization uncertainties. The mapping uncertainties can be, for example, multi-lane tracking uncertainties \cite{Yuxuan2024Fusion}, prediction uncertainties from vectorized map generators \cite{Qi2022ICRA-hdmapnet,Liu2023vectormapnet,liao2025maptrv2,yuan2024streammapnet} and image perception and calibration uncertainties \cite{Pengxin2024RAL}. The localization uncertainties arise mainly from the remaining positioning errors of the ego vehicles after pose graph optimization \cite{Onkar2017IROS,Christopher2020IV,Pengxin2024RAL}.
To address the uncertainties in map fusion, MapCVV \cite{Pengxin2024RAL} proposes an element-level optimization that can reduce  mapping and localization uncertainties of the lane lines at the same time. After optimization, a duplication removal module achieved by spatial depth-first search (DFS) is applied to obtain the final map.  

In this work, we represent 3D lane lines with B-splines, since their local control and flexibility make them well-suited to fuse uncertain, noisy crowd-sourced observations.  Note that, when a lane line is modeled as a sequence of Gaussian distributed B-spline control points, such a B-spline representation is not unique, which presents challenges when combining multiple B-splines into one. To this end, we propose a novel solution to fuse B-splines under different densities with uncertainties using pseudo measurements. We present a practical pipeline that uses pseudo measurements for grid search and association, followed by updates in the information form.

\begin{figure*}[th]
\centering \vspace{-4pt}
\includegraphics[width=0.75\linewidth]{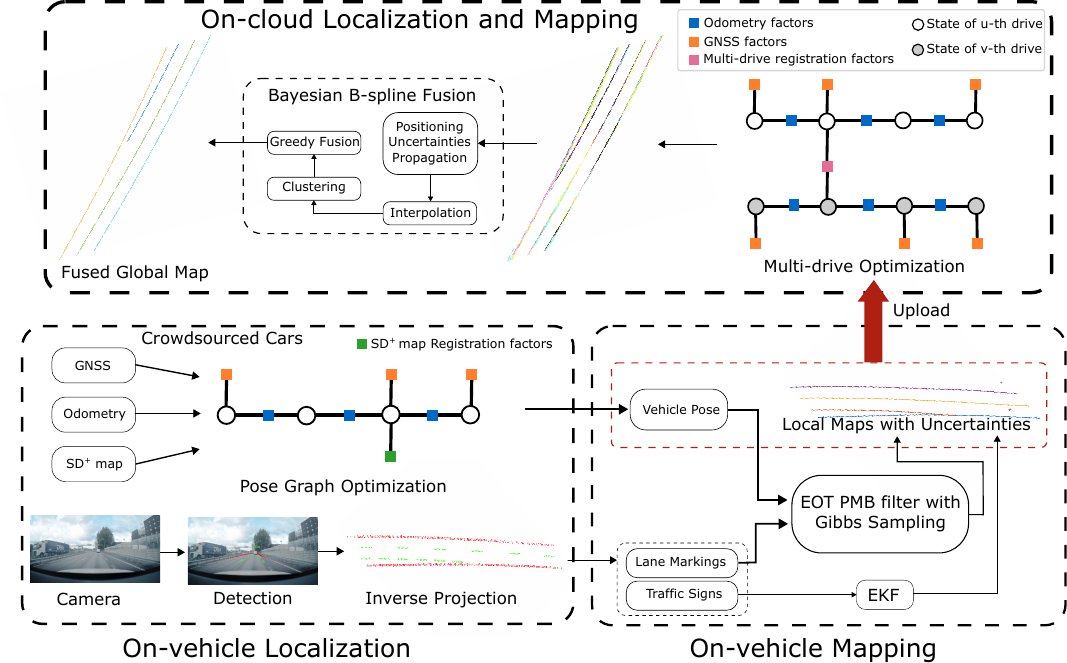}  
  \caption{System overview of B$^2$F-Map pipeline, including three modules: on-vehicle localization, on-vehicle mapping, and on-cloud localization and mapping. Note that traffic signs in the local maps are represented as semantic points and lane lines are represented by B-splines. With B-splines continuous over time, it is bandwidth efficient when uploading the control points to the cloud. On the cloud, after optimization to eliminate positioning errors in the estimated lane lines, the Bayesian B-spline fusion algorithm performs map fusion while maintaining the same B-spline representations.}
  \label{fig:system_overview}\vspace{-4pt}
\end{figure*}

In this work, we propose a pipeline that generates HD maps from scratch using production vehicles equipped with consumer-grade GNSS and IMU in a crowd-sourcing manner, shown in Fig.~\ref{fig:system_overview}. We name it \textbf{B$^2$F-Map} where B$^2$F refers to \textbf{B}ayesian \textbf{B}-spline \textbf{F}usion. Our contributions are:

\begin{itemize}
\item We propose a bandwidth-efficient crowd-sourced mapping pipeline consisting of on-vehicle localization, on-vehicle mapping with a Bayesian multi-lane tracker, and on-cloud localization and mapping, which includes multi-drive optimization and Bayesian map fusion. 
\item By representing the lane lines as 3D B-splines, we propose a novel approach to address the Bayesian fusion of B-spline trajectories with different densities. For lane-level map fusion, we discuss the handling of continuous and discontinuous partial overlaps. 
\item We validate the proposed approach on 70 km real-world data and release the dataset\footnote{https://github.com/yiping-xie/B2F-Map.}.
\end{itemize}

\section{Related Work}
\subsection{Local HD Mapping}
Various methods for building local maps have been proposed over the years. In the multi-stage end, most research relies on lane marking detection \cite{Onkar2017IROS} or segmentation \cite{Tong2021ICRA} for lane line mapping. The detected or segmented pixels can be projected to the world frame via inverse projection. In \cite{Onkar2017IROS}, a cubic spline is fitted locally to reconstruct a lane line.  In \cite{Tong2021ICRA}, a gridded semantic map is constructed after some noise filtering. Note that using a semantic map means that more storage space is needed on the vehicle and more bandwidth is required when uploading to the cloud. 

Recently, numerous online mapping frameworks \cite{Qi2022ICRA-hdmapnet,Liu2023vectormapnet,liao2023maptr} have been proposed to generate vectorized maps (mostly represented by polylines) directly from cameras and/or LiDAR on a frame-by-frame basis. The limitation is that they focus on per-frame map reconstruction, resulting in poor geometric consistency in the local maps from consecutive frames. Very recently, some works have attempted to address this by temporal fusion \cite{yuan2024streammapnet} or vectorized map tracking \cite{chen2024maptracker}. However, generating a temporally consistent HD map (longer than several hundred meters) in an end-to-end fashion, is still an open research question. 
Based on the per-frame online mapping \cite{liao2023maptr}, instead of building a local map on the vehicle, in \cite{Pengxin2024RAL}, all vectorized elements at sampled positions (at a 2-meter interval) are uploaded to the cloud. Then, a local map is generated on the cloud by B-spline fitting. However, the limitation is that since an element will be observed in multiple frames in the same drive, all replications will be uploaded to the cloud, which is far from bandwidth-efficient. 

Conceptually, the closest to the proposed approach is \cite{Yuxuan2024Fusion}, which frames the estimation of multiple lane lines as a multiple extended object tracking (EOT) problem, solved by a trajectory Poisson multi-Bernoulli mixture (TPMBM) filter. Such a formulation allows one to generate continuous lane lines (represented as a set of trajectories of B-spline control points) over time, which is promising for crowd-sourced mapping, considering the tight bandwidth constraint to upload data from customer vehicles to the cloud. Furthermore, the Bernoulli object spawning model in TPMBM filtering is suitable for modeling lane splitting. A key challenge in multiple extended object tracking is the data association problem, which, in the context of multi-lane tracking using point cloud measurements, refers to associating lane marking detection points to their corresponding lane lines correctly over time. In \cite{Yuxuan2024Fusion}, the data association problem is addressed by first clustering lane marking detection points into different groups and applying Murty's algorithm \cite{Murty} to only propagate global data association hypotheses with high weights over time.  
\subsection{Global HD Mapping}
In crowd-sourced mapping, the local maps constructed by different drives have unknown positioning errors, which need to be eliminated before fusing them into a globally consistent map. This problem is usually known as the multi-robot simultaneous localization and mapping (SLAM), commonly addressed using factor graph optimization \cite{Onkar2017IROS,Pannen2019IROS,Christopher2020IV,Paolo2022VTM}. A local map is typically divided into rigid submaps. A pose graph is then built, with submap poses as vertices and inter-drive loop closures (LCs) as edges. After using multi-drive graph optimization to reduce the unknown positioning errors, multiple observations of a map element (e.g., lane line) can be considered as globally consistent, in the sense that they can be assumed to be stacked on top of each other. To obtain the final map,  \cite{Onkar2017IROS} fits a spline to the sampled points. Similarly, \cite{Jian2023ITS} employs a gradual B-spline fitting algorithm after clustering.  \cite{Pengxin2024RAL} applies a DFS to remove duplications. \cite{Paolo2022VTM} applies a greedy pruning algorithm, which iteratively merges adjacent clothoids while maintaining continuity. 

\section{System Overview}
The proposed system consists of the following modules: (i) on-vehicle localization; (ii) on-vehicle mapping; (iii) on-cloud localization and mapping, as shown in Fig.~\ref{fig:system_overview}.
\subsection{On-Vehicle Localization} 
For the production vehicles, a graph optimization fusing GNSS, IMU, and lane marking registration with an $\textrm{SD}^+$ map is used for on-vehicle localization, as shown in Fig.~\ref{fig:system_overview}.  An $\textrm{SD}^+$ map is obtained from a standard-definition (SD) map with the estimation of the number of lanes and width of lanes from production vehicles. An SD link, given the lane count, can be transformed into ``HD'' lanes, where the lane width combined with the centerline is used to generate the position of lane markings. Although the generated $\textrm{SD}^+$ map has limited geometric accuracy at places like on-ramps, off-ramps, splits or merges, it can still be used as a backbone for positioning.  The detected lane markings, projected from image frame to world frame, are then used for iterative closest point (ICP) registration with $\textrm{SD}^+$ map. The optimized ego poses are used for the following on-vehicle mapping module.

\subsection{On-Vehicle Mapping}
Lane line mapping is formulated as a multi-lane tracking problem using an extended object trajectory (EOT) Poisson multi-Bernoulli (PMB) filter with Gibbs sampling. Traffic sign tracking employs an extended Kalman filter (EKF), as shown in Fig.~\ref{fig:system_overview}. We next outline the lane line state, multi-lane measurement, and dynamical models, and their role in recursive Bayesian estimation.
\subsubsection{Lane Line Modeling}
We model lane geometry using 3D quadratic B-splines of $d=2$, similar to \cite{Yuxuan2024Fusion}.
 A quadratic B-spline trajectory can be parameterized by $X=(\varepsilon, x^{1:v})$, where $\varepsilon$ is the initial time step\footnote{The initial time step is required to enable Bayesian filtering for sets of trajectories \cite{xia2023trajectory,Yuxuan2023Fusion-Gibbs}, but it is not used in modeling lane lines.} of the trajectory $X$, $v\geq3$ is its length, and $x^{1:v}=(x^1,\ldots,x^v)$, with $x^i \in \mathbb{R}^3$, denotes a finite sequence of control points. The continuous trajectory can be obtained by interpolating the control points using the B-spline basis function. For quadratic B-splines, each point on the continuous trajectory is determined by three consecutive control points. Specifically, the position of a point on the trajectory, determined by the subsequence of control points $x^{i:i+2}$, with $i\in\{1,\dots,v-2\}$, is \vspace{-4pt}
\begin{align}\label{eq:b_spline_interpolation}
  x(u) &= \Sigma(u)^T \otimes I_3 \times \begin{bmatrix}
    x^i \\ x^{i+1} \\ x^{i+2}
  \end{bmatrix}, \\
  \Sigma(u) &= \begin{bmatrix}
    1/2 & -1 & 1/2 \\
    1/2 & 1  & -1 \\
    0 & 0 & 1/2
  \end{bmatrix} \times \begin{bmatrix}
    1 \\ u \\ u^2
  \end{bmatrix},
\end{align}
where $u \in [0,1]$ and $I_3$ is an identity matrix. This means that $x(u)$ is a linear combination of control points $x^{i:i+2}$.
By moving $u$ from 0 to 1, the interpolated point $x(u)$ moves from the start position $x(0) = \frac{x^i+x^{i+1}}{2}$  to the end position $x(1) = \frac{x^{i+1}+x^{i+2}}{2}$ of the trajectory segment determined by control points $x^{i:i+2}$.
We assume that each control point of the B-spline trajectory $X$ is Gaussian distributed, i.e., 
\begin{equation}\label{eq:lane_line_model}
  p\left(x^i\right) = \mathcal{N}\left(x^i;m^i,P^i\right),
\end{equation} 
for $i\in \{1,\dots,v\}$. Then, it holds that every point on the B-spline trajectory is also Gaussian distributed. Specifically, assuming that there is no correlation between adjacent control point, the density of the interpolated point $x(u)$, determined by control points $x^{i:i+2}$, is given by 
\begin{align}
p\left(x(u)\right) &= \mathcal{N}\left(x(u);m(u),P(u)\right),\\
  m(u) &= H(u)\begin{bmatrix}
    m^i \;\;  m^{i+1} \;\; m^{i+2}
  \end{bmatrix}^T, \\
  P(u) &= H(u) \begin{bmatrix}
    P^i  \;\;  P^{i+1}  \;\;  P^{i+2}
  \end{bmatrix}^T,
\end{align}
where $H(u) = \Sigma(u)^T \otimes I_3$.

\subsubsection{Multi-Lane Measurement and Dynamic Model}\label{sec:multi_lane_measurement_and_dynamic_model}
For a trajectory $X=(\varepsilon, x^{1:v})$ at time step $k$, its interpolation at time step $k$ is determined by its latest three control points $x^{v-2:v}$.  Now we assume each individual lane marking edge detection point in the world frame $w_k$ follows Gaussian distribution, i.e., $\mathcal{N}(w_k;\varpi_k,\Omega_k^{w_k})$. Each measurement source $\varpi_k$ is considered uniformly distributed along the two lane marking edges, which is further approximated as a Gaussian  $\mathcal{N}(\varpi_k;h(\varepsilon,x^{\nu-2:\nu}),\Omega_k^\varpi)$. This modeling assumption gives the single measurement likelihood as
\begin{equation}
    \ell_k \left(w_k|\varepsilon,x^{\nu-2:\nu}\right) = \mathcal{N}\left(w_k;h\left(\varepsilon,x^{\nu-2:\nu}\right),\Omega_k^{w_k} + \Omega_k^\varpi\right),
\end{equation}
where the mean $h(\varepsilon,x^{\nu-2:\nu})$ is the interpolated point on trajectory $X$ at time step $k$, and can be computed using \eqref{eq:b_spline_interpolation}. The set of lane marking detections $\mathbf{w}_{k}$ generated by each lane line is modeled as a Poisson point process (PPP), parameterized by a Poisson rate $\lambda_k$, which models the average number of lane line detections. The complete multi-lane measurement likelihood equation is then 
\begin{equation}
    \ell_k(\mathbf z_k) = e^{-\lambda_k}\prod_{z_k \in \mathbf z_k}{\lambda_k \times \ell_k(w_k|\varepsilon,x^{v-2:v})}.
\end{equation}
For multi-lane dynamic model, it is the same as in  \cite{Yuxuan2024Fusion}.

\subsubsection{Bayesian Prediction and Update}
We recursively compute the posterior distribution of the set $\mathbf{X}_k$ of all B-spline trajectories given the measurements $\mathbf{w}_{1:k}$. The predicted density of the $\mathbf{X}_k$ at time step $k$ is
\begin{equation}\label{eq:multi_lane_prediction_step}
f\left(\mathbf{X}_k|\mathbf{w}_{1:k-1}\right) =  \int g_k\left(\mathbf{X}_k | \mathbf{X}_{k-1}\right) f\left(\mathbf{X}_{k-1}|\mathbf{w}_{1:k-1}\right) \delta \mathbf{X}_{k-1},
\end{equation}
where $g_k(\mathbf{X}_k | \mathbf{X}_{k-1}) $ is the transition density of the set of all trajectories for the multi-lane dynamic model, as described in \cite{Yuxuan2024Fusion}. %And the set integral $\int f(\mathbf{X}) \delta \mathbf{X}$ is defined in \cite{garcia2019multiple}.

The predicted density of the set of all B-spline trajectories at time step $k$ is then updated using the multi-lane measurement model $\ell_k(\mathbf{w}_k | \mathbf{X}_k)$ in Section \ref{sec:multi_lane_measurement_and_dynamic_model}, which gives
\begin{equation}\label{eq:multi_lane_update_step}
f\left(\mathbf{X}_k|\mathbf{w}_{1:k}\right) \propto f\left(\mathbf{X}_k|\mathbf{w}_{1:k-1}\right) \ell_k\left(\mathbf{w}_k | \mathbf{X}_k\right).
\end{equation}

The closed-form solution based on the above models is given by the extended object trajectory PMBM filter \cite{xia2023trajectory}. To perform the prediction and update steps in a computationally tractable way, one approach is to consider hard data associations between lane marking detection points \cite{Yuxuan2024Fusion}, i.e., use clustering information provided by the lane detector and assignment using Murty's algorithm. However, when lane marking detection points are very noisy and the clustering information brought by the lane detector is inaccurate, the data association solver using clustering and assignment may yield unreasonable data associations. Therefore, in this work, we adopt a more advanced multiple EOT algorithm, the TPMB filter using blocked Gibbs sampling \cite{Yuxuan2023Fusion-Gibbs}, which can yield soft assignments between lane marking detection points and B-spline trajectories.

\subsection{On-Cloud Localization and Mapping}
The next step is to fuse maps from multiple drives. To do this, the control points of B-splines generated from EOT PMB filter, the tracked traffic signs, and the ego vehicle poses are uploaded by production vehicles to the cloud for localization and mapping. This module consists of multi-drive optimization and Bayesian B-spline fusion. 
\subsubsection{Multi-Drive Optimization}
\begin{figure}[h]
\centering
\includegraphics[width=\linewidth]{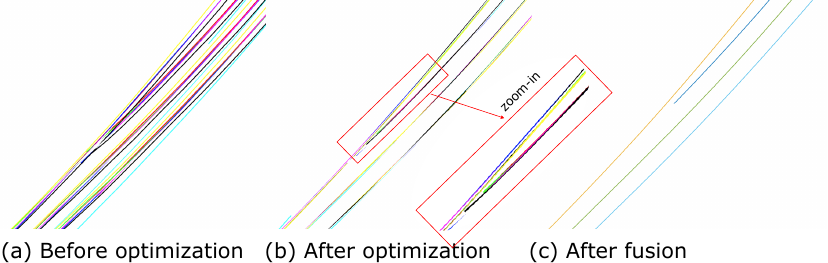}
  \caption{Visualization of a split/merge area. After optimization, positioning errors are reduced - lane lines observed multiple times are mostly stacked on top of each other. After fusion, redundant representation is removed.}
\label{fig:illustration_before_after_optimization_fusion}
\end{figure}

We consider $N$ drives indexed by $n \in \{1,\dots,N\}$, where each drive $n$ has discrete poses $\mathbf{T}_{n,k} \in \mathrm{SE}(3)$ at time step $k$, with
\begin{equation}
\mathbf{T}_{n,k}=\begin{bmatrix}\mathbf{R}_{n,k}&\mathbf{p}_{n,k}\\ \mathbf{0}^\top&1\end{bmatrix},\;\; 
\mathbf{R}_{n,k}\in\mathrm{SO}(3),\ \mathbf{p}_{n,k}\in\mathbb{R}^3.
\end{equation}
An illustration of the factor graph is shown in Fig.~\ref{fig:system_overview}, on-cloud localization module. The factor graph optimization can be formulated as follows:
\begin{equation}
\begin{aligned}
X^\star=\arg\min_{X}\ \sum_{n,k} \big\|\mathbf{r}_{n,k}^\text{o}\big\|^2_{\mathbf{\Omega}^\text{o}} + \sum_{g \in \mathcal{G}} \big\|\mathbf{r}_{n,k}^\text{g}\big\|^2_{\mathbf{\Omega}^\text{g}} \\ + \sum_{l \in \mathcal{L}} \sum_{\{u,v\}\in \mathcal{V}_l} \big\|\mathbf{r}_{uv,ij}^\text{LC}\big\|^2_{\mathbf{\Omega}^\text{r}} 
\end{aligned}
\end{equation}
where $X$ is the set containing all vehicle poses and the positions of all detected traffic signs, $\ell_m^{ts} \in \mathbb{R}^3$, namely,  $X=\{\{\mathbf{T}_{n,k}\}, \{\ell_m^{ts}\}_{m=1}^M\}$. Here, $\mathbf{r}_{n,k}^\text{o}$ is the odometry residual factor. $\mathbf{r}_{n,k}^\text{g}$ is the GNSS residual factor, and $\mathcal{G}$ is the set of the states that have good GNSS quality (e.g., those recorded outside tunnels). For a landmark $l \in \mathcal{L}$, which can be a submap of lane markings or a traffic sign, we denote $\mathcal{V}_l$ as the set containing all the drives that pass this submap.  $\mathbf{r}_{uv,ij}^\text{LC}$ is the registration residual between drive $u$ and drive $v$ from inter-drive loop closures, where $i$ and $j$ are time steps of the submap center of drive $u$ and $v$, respectively. 
Note that we use the notation $\big\|\mathbf{r}\big\|^2_{\mathbf{\Omega}}:=\mathbf{r}^T\mathbf{\Omega}^{-1}\mathbf{r}$ to denote the squared Mahalanobis distance. For matching lane lines represented by B-splines, we first sample a point cloud from the control points and use semantic ICP \cite{gong2024lidar} for registration. Fig.~\ref{fig:illustration_before_after_optimization_fusion} shows the changes of a split/merge area (a) before and (b) after multi-drive optimization.
\subsubsection{Bayesian Map Fusion}
Map fusion takes a set of lane lines and fuses them into a globally consistent lane-level map without redundant representations. As inputs, each lane line is modeled using a sequence of B-spline control points and each control point has a Gaussian density distribution. The detailed process is described in the following section.

\section{Bayesian B-splines Fusion}\label{sec:b-spline_fusion}
In this section, we describe (i) how to estimate and propagate lane line positioning uncertainties, then (ii) how to perform Bayesian lane-level map fusion with B-splines. Section~\ref{sec:fusion_two_overlapping} covers the fusion of two overlapping lane lines, Section~\ref{sec:fusion_two_partially_overlapping} generalizes this to partial overlaps, and finally, Section~\ref{sec:fusion_multiple} presents a greedy algorithm for fusing sets of multiple lane lines.

\subsection{Propagating Positioning Uncertainty}
After multi-drive optimization, for the same element (e.g., a lane line), observations from multiple drives should ideally be very close to each other. However, there usually exists some residual positioning errors, shown in the zoom-in window in Fig.~\ref{fig:illustration_before_after_optimization_fusion}(b). To address this, we use relative errors across different drives \cite{Pengxin2024RAL} denoted as $\epsilon_l^n$, as an indication of the positioning uncertainties and propagate their absolute values, $\big\|\mathbf{\epsilon}_l^n\big\|$ to the B-splines, such that, $P^i$ in~\eqref{eq:lane_line_model} becomes $P^i \big\|\mathbf{\epsilon}_l^n\big\|$, which will be used in multi-lane fusion. Here, the relative error $\mathbf{\epsilon}_l^n$ is defined as:
\begin{align}
    \bar{\mathbf{v}}_l&=\mathbf{SVD}\left( \{ \mathbf{v}_l^n \} |_{n=1,2,\ldots,N} \right), \\
    \mathbf{\epsilon}_l^n &= \mathbf{v}_l^n \boxminus \bar{\mathbf{v}}_l. \label{eq:global consistency}
\end{align}
Specifically, for the $l^{th}$ element, we use singular value decomposition (SVD) on different drives to compute the implicit vector element $\bar{\mathbf{v}}_l$. Then, for the $l^{th}$ element from the $n^{th}$ drive, we calculate its relative error $\mathbf{\epsilon}_l^n$ as the geometric discrepancy between $\bar{\mathbf{v}}_l$ and $\mathbf{v}_l^n$. In~\eqref{eq:global consistency}, we use $\boxminus$ to denote the point-to-line distance for lane lines and point-to-point distance for traffic signs. 

\begin{algorithm}[!t]\scriptsize
  \caption{Fusion of two B-splines $X_1$ and $X_2$ (fusing B-spline trajectory $X_2$ into $X_1$)}
  \begin{algorithmic}[1]
      \STATE \textbf{Input:} B-spline control point sequences $X_1 = x_1^{1:v_1}$ with means $(m_1^1,\dots,m_1^{v_1})$ and covariances $(P_1^1,\dots,P_1^{v_1})$ and $X_2 = x_2^{1:v_2}$ with means $(m_2^1,\dots,m_2^{v_2})$ and covariances $(P_2^1,\dots,P_2^{v_2})$, $M$, and $\tau$.
      \STATE \textbf{Output:} Fused B-spline control point sequence with means $(m_f^1,\dots,m_f^{v_1})$ and covariances $(P_f^1,\dots,P_f^{v_1})$.
      \STATE Initialize the sequence of pseudo measurements $(z^1_1,\dots,z^1_M,\dots,z^{v_2-2}_1,\dots,z^{v_2-2}_M)$ as interpolated points of $X_2$.\label{alg_line:pseudo_measurements_start}
      \FOR{$i = 1$ \textbf{to} $v_2-2$}
          \FOR{$j = 1$ \textbf{to} $M$}
              \STATE $z^i_j = H((j-1)/M)\mathbf{m}_2^i$
          \ENDFOR
      \ENDFOR \label{alg_line:pseudo_measurements_end}
      \STATE Initialize the sequence of interpolated points $(x^1_1,\dots,x^1_\tau,\dots,x^{v_1-2}_1,\dots,x^{v_1-2}_\tau)$ of $X_1$.\label{alg_line:interpolated_points_start}
      \FOR{$i = 1$ \textbf{to} $v_1-2$}
          \FOR{$t = 1$ \textbf{to} $\tau$}
              \STATE $x^i_t = H((t-1)/\tau)\mathbf{m}_1^i$
          \ENDFOR
      \ENDFOR \label{alg_line:interpolated_points_end}
      \STATE Initialize $(u^1_1,\dots,u^1_M,\dots,u^{v_2-2}_1,\dots,u^{v_2-2}_M)$.\label{alg_line:grid_search_start}
      \STATE Initialize $(i^1_1,\dots,i^1_M,\dots,i^{v_2-2}_1,\dots,i^{v_2-2}_M)$.
      \FOR{each element $z_j^{i^\prime}$ in $(z^1_1,\dots,z^1_M,\dots,z^{v_2-2}_1,\dots,z^{v_2-2}_M)$}
        \FOR{each element $x_t^{\iota}$ in $(x^1_1,\dots,x^1_\tau,\dots,x^{v_1-2}_1,\dots,x^{v_1-2}_\tau)$}
          \STATE compute $d^\iota_t = \|z^{{i^\prime}}_j - x_t^{\iota}\|_2$.
        \ENDFOR
        \STATE Find the minimum $d^{\iota^\prime}_{t^\prime}$, set $u^{{i^\prime}}_j = (t^\prime-1)/\tau$,  $i^{{i^\prime}}_j = \iota^\prime$.
      \ENDFOR \label{alg_line:grid_search_end}
      \STATE Set $(m_f^1,\dots,m_f^{v_1})=(m_1^1,\dots,m_1^{v_1})$ and $(P_f^1,\dots,P_f^{v_1})=(P_1^1,\dots,P_1^{v_1})$. \label{alg_line:information_update_start}
      \FOR{$i = 1$ \textbf{to} $v_1-2$}
        \STATE Compute information vector and matrix of $\mathbf{m}_f^i$ and $\mathbf{P}_f^i$ using \eqref{eq:information_vector} and \eqref{eq:information_matrix}.
        \STATE Find all $i^{{i^\prime}}_j = i$ and their corresponding $u^{{i^\prime}}_j$ and $z_j^{i^\prime}$.
        \STATE Perform information update using \eqref{eq:update_information_vector} and \eqref{eq:update_information_matrix}.
        \STATE Recover the updated mean $\mathbf{m}_f^i$ and covariance $\mathbf{P}_f^i$ using \eqref{eq:update_mean} and \eqref{eq:update_covariance}. 
      \ENDFOR \label{alg_line:information_update_end}
  \end{algorithmic}
  \label{algorithm_Bspline_fusion}
\end{algorithm}
\subsection{Fusion of Two Overlapping Lane Lines with Pseudo Measurements}\label{sec:fusion_two_overlapping}
We model a continuous lane line using a sequence of B-spline control points. The same lane can be represented by different control point sequences, possibly with different numbers of control points, leading to varying density representations. To address this, we fuse two B-splines using interpolated points as pseudo measurements, as follows.
\subsubsection{Information Update}
To fuse the lane line geometry and uncertainty information captured by two overlapping B-spline trajectories with different densities, we first interpolate one B-spline trajectory to obtain a sequence of interpolated points, which is then used as pseudo measurements $z\in\mathbb{R}^3$ to update the other B-spline trajectory. 
Suppose that we have $M$ estimates of a sequence of interpolated points $(x(u_1),\ldots,x(u_M))$, all determined by the control points $x^{i:i+2}$, and they have a sequence of pseudo measurements $(z_1,\ldots,z_M)$ with noise covariances $(R_1,\ldots,R_M)$.
We use the information form to perform the update, where the mean and covariance are replaced by information vector and information matrix, respectively. We define $\mathbf{m}^i=\textrm{vec}(m^i,m^{i+1},m^{i+2})$ as the vectorization of control point sequence $(m^i,m^{i+1},m^{i+2})$ and $\mathbf{P}^i=\textrm{diag}(P^i, P^{i+1}, P^{i+2})$, their corresponding information vector and information matrix are
\begin{subequations} \allowdisplaybreaks
\begin{align}
    \mathbf{y}^i&=(\mathbf{P}^i)^{-1}\mathbf{m}^i \label{eq:information_vector} \\
    \mathbf{Y}^i&=(\mathbf{P}^i)^{-1}. \label{eq:information_matrix}
\end{align}
\end{subequations}
The information update of $\mathbf{y}^i$ and $\mathbf{Y}^i$ using pseudo measurements is then given by:
\begin{subequations}\allowdisplaybreaks
\begin{align}
    \mathbf{y}_f &= \mathbf{y}^i + \frac{1}{M}\sum_{j=1}^M H(u_j)^TR_j^{-1}z_j, \label{eq:update_information_vector}\\
    \mathbf{Y}_f &= \mathbf{Y}^i + \frac{1}{M}\sum_{j=1}^M H(u_j)^TR_j^{-1}H(u_j), \label{eq:update_information_matrix}
\end{align}
\end{subequations}
where $\mathbf{y}_f$ and $ \mathbf{Y}_f$ are the updated information vector and information matrix, respectively. The mean and covariance of the updated control point sequence $x^{i:i+2}_f$ can be recovered from $\mathbf{y}_f$ and $\mathbf{Y}_f$ via 
\begin{subequations}\allowdisplaybreaks
\begin{align}
\mathbf{m}_f&=\mathbf{Y}_f^{-1}\mathbf{y}_f \label{eq:update_mean}\\
    \mathbf{P}_f&=\mathbf{Y}_f^{-1}. \label{eq:update_covariance}
\end{align}
\end{subequations}
The advantage of the information update is that multiple measurements can be filtered simultaneously simply by summing their corresponding information vectors and matrices. 

\subsubsection{Grid search}
Note that the above formulation is based on the assumption that, for each pseudo measurement $z_j$ on one B-spline $X=x^{1:v}$, we know the interpolated point $x(u_j)$ to which it corresponds to on the other B-spline. To find $u_j$ and the subsequence of control points $x^{i:i+2}$ where $i\in\{1,\ldots,v-2\}$, given $z_j,j\in\{1,\ldots,J\}$, from a sequence of pseudo measurements $(z_1,\ldots,z_J)$, we can formulate the solution as solving the following bivariate optimization:
\begin{equation}
    \min_{u_j,i}\big\| z_j - H(u_j)\mathbf{m}^i\big\|_2, \label{eq:optimization_grid_search}
\end{equation}
which minimizes the Euclidean distance between $z_j$ and $x(u_j)$ determined by $u_j$ and the means of the control points $x^{i:i+2}$. Instead of solving such a complex optimization problem for every pseudo measurement, we adopt a simpler solution based on grid search. More importantly, the grid search also enables us to easily identify the parts of the two partially overlapping lane lines that need to be fused. 

The pseudo code for fusing two overlapping B-spline trajectories is given in Algorithm \ref{algorithm_Bspline_fusion}. Lines \ref{alg_line:pseudo_measurements_start}-\ref{alg_line:pseudo_measurements_end} describe how to obtain pseudo measurements from $X_2$; lines \ref{alg_line:interpolated_points_start}-\ref{alg_line:interpolated_points_end} precompute the interpolated points on $X_1$, for the grid search later (lines \ref{alg_line:grid_search_start}-\ref{alg_line:grid_search_end}); and lines \ref{alg_line:information_update_start}-\ref{alg_line:information_update_end} present the information update.

\subsection{Fusion of Two Partially Overlapping Lane Lines}\label{sec:fusion_two_partially_overlapping}
 In practice, two lane line estimates, representing part of the same lane line obtained using fleet data collected on different routes, only partially overlap with each other. In these cases, we only need to fuse the subsequences of B-spline control points that represent the same segment(s) of the lane line. Hence, we need to identify, for each B-spline trajectory, the subsequence of control points that represents the overlapping area. To achieve this, we first compare the minimum value of~\eqref{eq:optimization_grid_search} and compare it with a pre-defined threshold $\Gamma$. If there are at least $\tau$ consecutive interpolated points $x(u_j)$ of $X_1$ whose minimum value are smaller that $\Gamma$, then these two lane lines are considered to be partially overlapping. For two lane lines that are partially overlapping, the overlapping area can either be continuous (\textit{Case 1-4}) or discontinuous (\textit{Case 5}). In the following, we will discuss solutions in different situations. Suppose that we have two B-spline trajectories $X_1=(x_1^1,\ldots,x_1^{v_1})$ and $X_2=(x_2^1,\ldots,x_2^{v_2})$.
\begin{figure}[!ht]
\centering
\includegraphics[width=\linewidth]{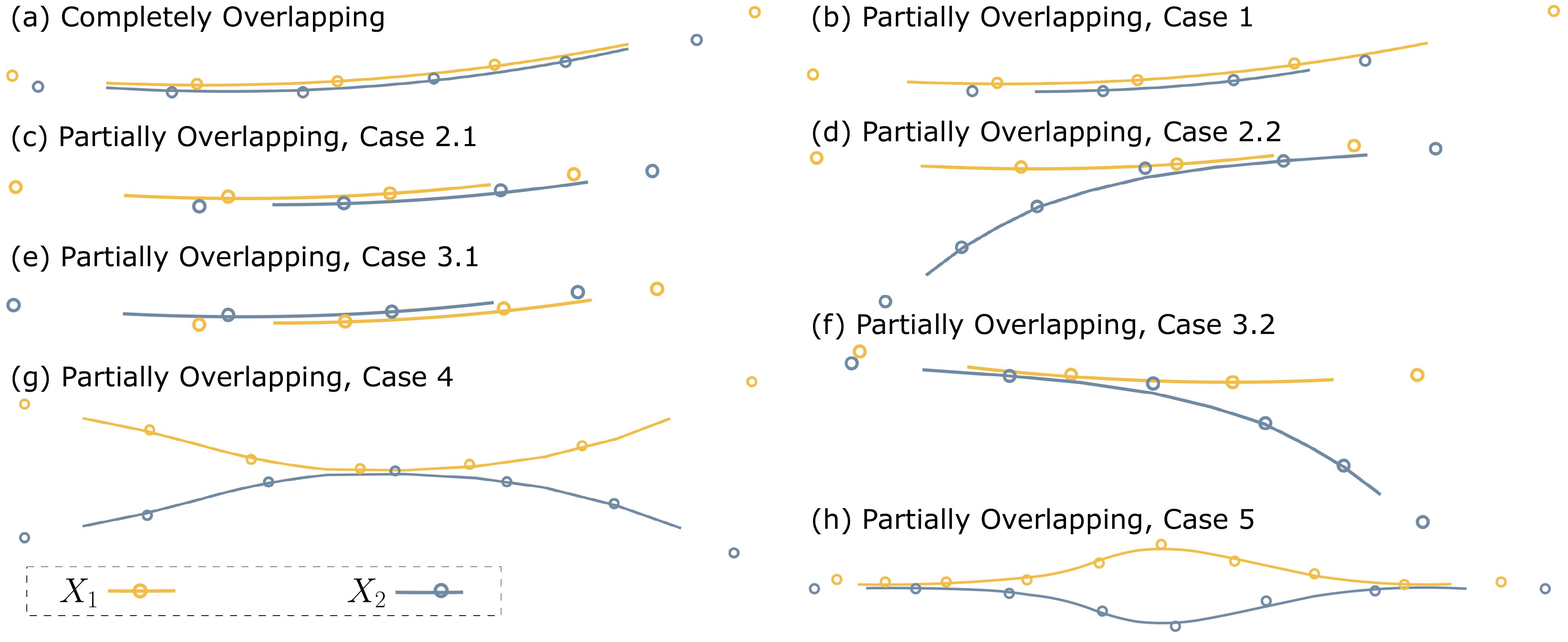}
  \caption{Illustration of two B-splines, completely overlapping (closely-spaced) in (a), partially overlapping in (b)-(h). \textit{Case 1-5} are presented in details in Section~\ref{sec:fusion_two_partially_overlapping}. Note that (c) and (d) both correspond to \textit{Case 2}, where in (c) the two trajectories would be merged into one whereas in (d), the trajectory in blue will be truncated into two parts. Similar for \textit{Case 3} in (e) and (f). (g) illustrates \textit{Case 4} where an interior subsequence of one B-spline overlaps with a subsequence of another. (h) illustrates the case where the overlapping area is discontinuous, e.g., when traffic islands are present.}
\label{fig:fusion_all_cases}
\end{figure}
\subsubsection*{Case 1, one B-spline completely overlaps with a portion of another}
Without loss of generality, we assume that $X_2$ overlaps with the subsequence of control points $x_1^{i:j}$ of $X_1$, as shown in Fig.~\ref{fig:fusion_all_cases}(b). Then we can interpolate $X_2$ into pseudo measurements and use them to update $x_1^{i:j}$ to obtain the fused control points $x_f^{1:j-i+1}$. After fusion, $X_1$ and $X_2$ are merged into a single B-spline trajectory, given by $X_f=(x_1^1,\ldots,x_1^{i-1},x_f^1,\ldots,x_f^{j-i+1},x_1^{j+1},\ldots,x_1^{v_1})$.
\subsubsection*{Case 2, the beginning of one B-spline overlaps with a subsequence of another}
Assume that a subsequence of control points $x_1^{1:\iota}$ of $X_1$ including $x_1^1$ overlaps with a subsequence of control points $x_2^{i:j}$ of $X_2$. To fuse $X_1$ and $X_2$, we first truncate $X_2$ into at most three parts: $x_2^{1:i-1}$, $x_2^{i:j}$ and $x_2^{j+1:v_2}$. Then we interpolate $x_2^{i:j}$ into pseudo measurements and use them to update $x_1^{1:\iota}$ to obtain the fused control points $x_f^{1:\iota}$. After fusion, we obtain B-spline trajectory $(x_2^1,\ldots,x_2^i,x_f^1,\ldots,x_f^\iota,x_1^{\iota+1},\ldots,x_1^{v_1})$, which concatenates the first part of $X_2$, the fused control points and the rest of the control points of $X_1$, and also truncated B-spline trajectory $x_2^{j+1:v_2}$. Note that when $j=v_2$, B-spline trajectory $x_2^{j+1:v_2}$ does not exist, and the two lane lines are joined into a longer lane line, which is the case illustrated in Fig.~\ref{fig:fusion_all_cases}(c). 

In scenarios with merging and splitting lanes [see Fig.~\ref{fig:fusion_all_cases}(d)], we could have $j<v_2$. To make $x_2^{j+1:v_2}$ a valid quadratic B-spline trajectory with at least 3 control points, we append it to let it become $(x_f^{\iota-1},x_f^{\iota},x_2^{j+1},\ldots,x_2^{v_2})$ with two shared control points of the fused B-spline trajectory. This also makes sure that the two continuous lane lines obtained by interpolating $(x_f^{\iota-1},x_f^{\iota},x_2^{j+1},\ldots,x_2^{v_2})$ and the fused B-spline trajectory have at least one point in common, which well models lane split/merge.

\subsubsection*{Case 3, the end of one B-spline overlaps with a subsequence of another}
Assume that a subsequence of control points $x_1^{\iota:v_1}$ of $X_1$ including $x_1^{v_1}$ overlaps with a subsequence of control points $x_2^{i:j}$ of $X_2$. Similar to \textit{Case 2}, we also first truncate $X_2$ into at most three parts, and then we consider $x_2^{i:j}$ as pseudo measurements to update $x_1^{\iota:v_1}$ to obtain the fused control points $x_f^{1:v_1-\iota+1}$. After fusion, we obtain B-spline trajectory $(x_1^1,\ldots,x_1^{\iota-1},x_f^1,\ldots,x_f^{v_1-\iota+1},x_2^{j+1},\ldots,x_2^{v_2})$, which concatenates the first part of $X_1$, the fused control points and the rest of the control points of $X_2$, and also truncated B-spline trajectory $x_2^{1:i-1}$. Also, similar to \textit{Case 2}, when $i>1$ [shown in Fig.~\ref{fig:fusion_all_cases}(e)], we append the truncated B-spline trajectory to let it become $(x_2^1,\ldots,x_2^{i-1},x_f^1,x_f^2)$.

\subsubsection*{Case 4, an interior subsequence of one B-spline overlaps with a subsequence of another}
Assume that a subsequence of control points $x_1^{\iota:\iota'}$ of $X_1$, where $\iota>1$ and $\iota'<v_1$, overlaps with a subsequence of control points $x_2^{i:j}$ of $X_2$, illustrated in Fig.~\ref{fig:fusion_all_cases}(g). Similar to \textit{Case 1} and \textit{Case 2}, we first truncate $X_2$ into at most three parts, and then we consider $x_2^{i:j}$ as pseudo measurements to update $x_1^{\iota:{\iota}'}$ to obtain the fused control points $x_f^{1:\iota'-\iota+1}$. After fusion, we obtain B-spline trajectory $(x_1^1,\ldots,x_1^{\iota-1},x_f^1,\ldots,x_f^{\iota'-\iota+1},x_1^{\iota'+1},\ldots,x_1^{v_1})$, which replaces $x_1^{\iota:\iota'}$ in $X_1$ with $x_f^{1:\iota'-\iota+1}$, and two truncated B-spline trajectories $x_2^{1:i-1}$ and $x_2^{j+1:v_2}$. Finally, we append the two truncated B-spline trajectories to let them become $(x_2^1,\ldots,x_2^{i-1},x_f^1,x_f^2)$ and $(x_f^{\iota'-\iota},x_f^{\iota'-\iota+1},x_2^{j+1},\ldots,x_2^{v_2})$.

\subsubsection*{Case 5, discontinuous overlapping area} The overlapping area can be discontinuous, for example, when traffic islands are present, as lane lines may first split and then merge to direct traffic flow. In these cases, it is possible that the two lane line estimates only overlap before and after the traffic island, as shown in Fig.~\ref{fig:fusion_all_cases}(h). In practice, to detect a discontinuous overlapping area, we can check if there exists more than one sequence of interpolated points in which there are at least $\tau$ consecutive interpolated points $x(u_j)$ of $X_1$ whose minimum value of~\eqref{eq:optimization_grid_search} are smaller than $\Gamma$. If so, we truncate the shorter B-spline trajectory into different parts at control points corresponding to the boundaries of overlapping areas. By doing so, we only have lane lines with continuous overlapping areas.  

We note that, in cases where the two B-spline trajectories are not merged into a single one, we need to truncate one of the B-spline trajectories into different parts and concatenate the fused B-spline trajectory with the truncated ones. It is clear that such an operation is not unique, similar to how the representation of merge or split lane lines is also not unique. In the fusion process introduced above, we choose to truncate the B-spline trajectory that has been interpolated into pseudo measurements while increasing the length of the other B-spline trajectory by concatenation.

\subsection{Fusion of Multiple Lane Lines}\label{sec:fusion_multiple}
To perform lane-level map fusion, we extend the fusion method for merging two partially overlapping lane lines to the fusion of multiple lane lines. The challenge of multi-lane fusion is that, given two sets of B-spline trajectories, the mapping between these two sets is not injective nor surjective, as multiple lane lines from one set can be fused with a single lane line from the other set, and vice versa. Moreover, the vehicle could revisit the road it has traveled, and thus the set of lane line estimates obtained in a single drive may already contain overlapping lane line estimates.

To solve this problem, we adopt a greedy fusion algorithm that takes a set of B-spline trajectories as input and outputs a fused set. The idea is simple: we group all the sets of B-spline trajectories, with each set obtained from a single drive, into a sequence of B-spline trajectories (in arbitrary order). Then for each trajectory of the sequence, we check if it can be fused with another trajectory in the sequence sequentially. If fusion can be performed, then we fuse these two trajectories, and based on the fusion result, the sequence of trajectories may need to be extended due to truncation of existing trajectories. Also note that a fused trajectory can still be fused with other trajectories. However, directly solving this problem can be computationally demanding. A more efficient implementation can be obtained by using clustering. First, we interpolate all the B-spline trajectories, and we apply clustering to all these interpolated points. Based on the clustering results, we group trajectories in the sense that trajectories in different groups should not have any interpolated points within the same cluster. Finally, we can perform multiple lane lines fusion within each individual group.
\section{Experiments}
We validate the proposed pipeline on real-world datasets collected using production vehicles in two cities in Europe. The datasets contain two areas, Area 1 and Area 2, covering highways and tertiary roads under varied lightning and weather, totaling 70 km. Each area includes 8 drives, with vehicles factory-calibrated  offline with an additional real-time dynamic extrinsic calibration. Each vehicle is equipped with a separate high-precision localization system from Oxford Technical Solutions (OxTS), providing ground-truth positioning. As for HD maps, we use data provided by professional surveying vehicles as ground truth. 

We conduct two experiments to evaluate the performance of the proposed approach. The first experiment aims to evaluate the map quality of the on-vehicle mapping module, where we choose TPMBM filter with clustering-based DA \cite{Yuxuan2024Fusion} as the baseline. The second one benchmarks the absolute accuracy and relative accuracy of the final HD map produced by B$^2$F-Map pipeline. In this experiment, for the baseline, we keep the on-vehicle localization, on-vehicle mapping and multi-drive optimization modules the same, but only replace the Bayesian B-spline fusion module with cubic splines that fit the sampled points of all observed lane lines within a cluster \cite{Onkar2017IROS}. 
\subsection{Multi-Lane Tracking Performance}
\begin{figure}[!h]
\centering
\includegraphics[width=\linewidth]{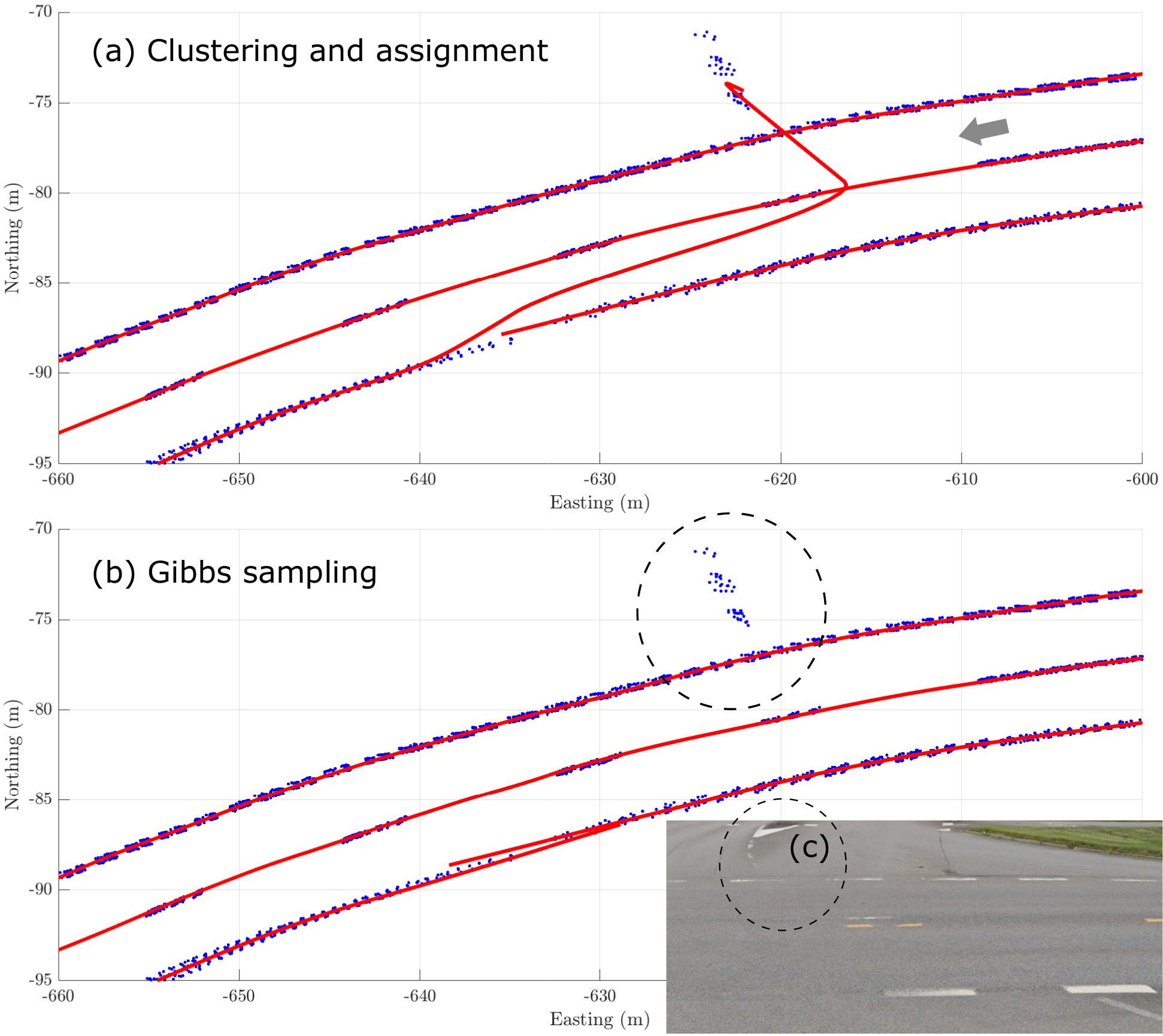}
  \caption{Qualitative comparisons of clustering and assignment DA using Murty's algorithm \cite{Yuxuan2024Fusion} (baseline) in (a) and Gibbs sampling-based DA in (b). The blue points are the noisy lane marking edge detections and the red lines are lane lines produced by EOT tracker. The gray arrow in (a) illustrates the vehicle's travel direction. (c) shows the Google street view of the lane markings from another lane line, which are wrongly associated in baseline approach, but not in the Gibbs sampling based approach.}
\label{fig:result_pmbm_oxts}
\end{figure}

A problematic case of multi-lane tracking with clustering-based DA is shown in Fig.~\ref{fig:result_pmbm_oxts}(a), where during tracking, the lane marking detections from another (almost perpendicular) lane line are wrongly associated to a lane line parallel to the vehicle's travel direction. However, using Gibbs sampling-based DA in our proposed approach avoids this mistake, shown in Fig.~\ref{fig:result_pmbm_oxts}(b). Table~\ref{tab:DA_compare} shows the number of wrong associations for all drives in both areas.  The results show that in almost all drives, the Gibbs sampling–based approach outperforms the baseline, specially with large improvements on the challenging cases such as Drive 04, 06 and 07 from Area 1. This experiment demonstrates the advantages of robustness of our proposed on-vehicle mapping algorithm over baseline.  
\vspace{0pt}

\begin{table}[h]
\centering
\caption{No. of Wrong associations}\label{tab:DA_compare}
\begin{tabular}{c|cc|cc}
\cline{1-5}
\multirow{2}{*}{Drive} & \multicolumn{2}{c}{Area 1} & \multicolumn{2}{c}{Area 2} \\ \cline{2-5} 
                      & Clustering      & Gibbs sampling      & Clustering      & Gibbs sampling      \\ \cline{1-5}
01                    & 3             & 0         &  2             &     0      \\ 
02                    & 3             & 1         &  3             &     0      \\
03                    & 5             & 1        &   3            &     0      \\
04                    & 6             & 2        &  4             &     0      \\
05                    & 4             & 2         &  3             &    1       \\
06                    & 7             & 1         & 1              &    1       \\
07                    & 8             & 3         &  3             &    1       \\
08                    & 4             & 2         &  4             &    0      \\ \cline{1-5}
\end{tabular}\vspace{0pt}
\end{table}

\subsection{Lane Lines Absolute and Relative Accuracy Benchmark}
Tables~\ref{tab:abs_error} and~\ref{tab:rel_error} show the mean $\mu$ and standard deviation $\sigma$ of absolute errors and relative errors of crowd-sourced maps generated using B$^2$F-Map pipeline. We can see that the relative errors are much smaller than absolute errors, indicating that the produced crowd-sourced maps are geometrically accurate but the lane lines can have some offsets compared to the ground truth.  However, in the autonomous driving industry, HD maps are widely used for localization and downstream tasks such as trajectory planning, where relative accuracy is typically more important.

\begin{figure}[!h]
\centering
\includegraphics[width=\linewidth]{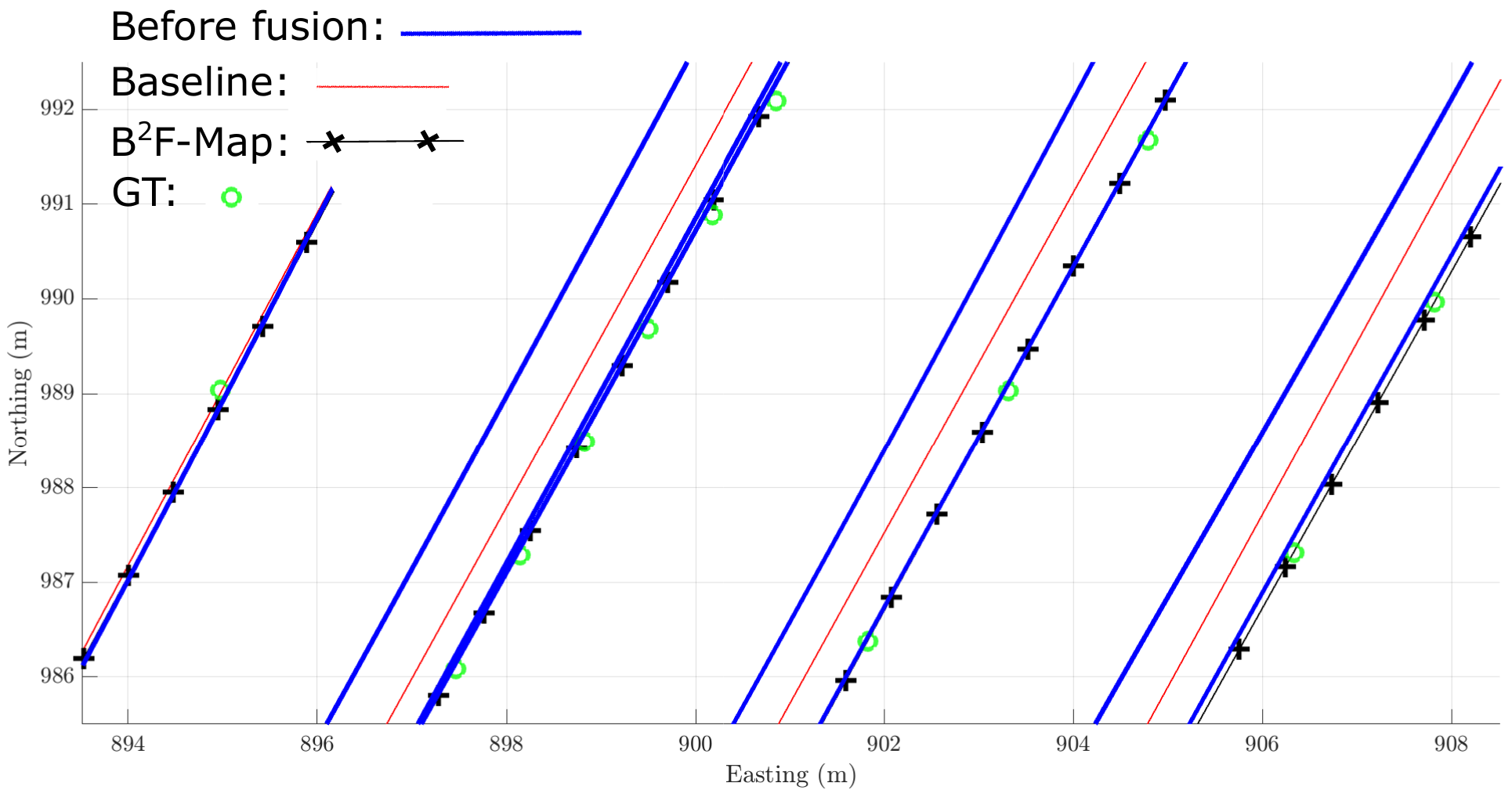}
  \caption{A qualitative result on the crowd-sourced map, where the fused lane lines from B$^2$F-Map pipeline are in black, and the ones from baseline are in red. The estimated lane lines from crowd-sourced vehicles before fusion are in blue and the ground truth is in green. Note that B$^2$F-Map results in accurate estimates despite the errors in some lane lines before fusion. }
\label{fig:result_fusion_compare}
\end{figure}

For ablation study on the Bayesian B-spline fusion module, we compare our approach to the baseline, which fuses lane lines from different drives by clustering lane lines and then fitting the sampled points within a cluster \cite{Onkar2017IROS}. Note that the main issue of the baseline is that it only fuses the geometry of the lane lines but ignores the positioning and mapping uncertainties of the lane lines while our B$^2$F module fully utilizes the uncertainties during fusion. We showcase the advantages of our approach in Fig.~\ref{fig:result_fusion_compare}, where after multi-drive optimization, some estimated lane lines still suffer from large positioning and mapping errors before fusion. We can see that compared to the baseline, our approach results in much more accurate estimates after fusion in terms of both absolute accuracy and lane width. As a result, in Area 1, which is the more challenging dataset, we can see from Table~\ref{tab:abs_error} and Table~\ref{tab:rel_error}, our approach improves a lot compared to the baseline.

\vspace{0pt}

\vspace{0pt}
\begin{table}[h]
\centering
\caption{Absolute error on lane lines (unit: meter)}\label{tab:abs_error}
\begin{tabular}{c|cc|cc}
\cline{1-5}
\multirow{2}{*}{Method} & \multicolumn{2}{c}{Area 1} & \multicolumn{2}{c}{Area 2} \\ \cline{2-5} 
                      & $\mu$      & $\sigma$      & $\mu$      & $\sigma$      \\ \cline{1-5}
Baseline                    & 0.612             & 0.351         & 0.772            &     \textbf{0.420}      \\ 
B$^2$F-Map                   & \textbf{0.585}             & \textbf{0.333}        &  \textbf{0.760}             &     0.422      \\
\cline{1-5}
\end{tabular}
\end{table}\vspace{0pt}

\begin{table}[h]
\centering
\caption{Relative error on lane lines (unit: meter)}\label{tab:rel_error}
\begin{tabular}{c|cc|cc}
\cline{1-5}
\multirow{2}{*}{Method} & \multicolumn{2}{c}{Area 1} & \multicolumn{2}{c}{Area 2} \\ \cline{2-5} 
                      & $\mu$      & $\sigma$      & $\mu$      & $\sigma$      \\ \cline{1-5}
Baseline                    & 0.133           & 0.126         & \textbf{0.071}            &     \textbf{0.055}      \\ 
B$^2$F-Map                   & \textbf{0.117}           & \textbf{0.108}        &  0.079            &     0.060      \\
\cline{1-5}
\end{tabular}
\end{table}

\section{CONCLUSIONS}
In this work, we propose a crowd-sourced mapping pipeline, B$^2$F-Map, without relying on a base HD map. We use B-splines for lane line representations throughout the whole pipeline, where the on-vehicle mapping module adapts the current state-of-the-art multiple EOT algorithm, TPMB filter with Gibbs sampling, to ensure robust data association. Whereas for on-cloud mapping, we propose a novel Bayesian B-spline fusion algorithm to fuse lane-level maps from crowd-sourced maps utilizing both geometry and uncertainty,  effectively fusing B-spline trajectories under different densities. The experiments on real-world datasets demonstrate that the proposed approach is capable of producing high-quality HD maps in a crowd-sourcing manner.

The current major limitation of B$^2$F-Map is that it does not contain lane topology, which we intend to address in the future. Future work also includes improving the real-time performance of the on-vehicle mapping module.  
\addtolength{\textheight}{-12cm}   % This command serves to balance the column lengths
                                  % on the last page of the document manually. It shortens
                                  % the textheight of the last page by a suitable amount.
                                  % This command does not take effect until the next page
                                  % so it should come on the page before the last. Make
                                  % sure that you do not shorten the textheight too much.

%%%%%%%%%%%%%%%%%%%%%%%%%%%%%%%%%%%%%%%%%%%%%%%%%%%%%%%%%%%%%%%%%%%%%%%%%%%%%%%%

%%%%%%%%%%%%%%%%%%%%%%%%%%%%%%%%%%%%%%%%%%%%%%%%%%%%%%%%%%%%%%%%%%%%%%%%%%%%%%%%

%%%%%%%%%%%%%%%%%%%%%%%%%%%%%%%%%%%%%%%%%%%%%%%%%%%%%%%%%%%%%%%%%%%%%%%%%%%%%%%%
% \section*{APPENDIX}

% Appendixes should appear before the acknowledgment.

% \section*{ACKNOWLEDGMENT}

%%%%%%%%%%%%%%%%%%%%%%%%%%%%%%%%%%%%%%%%%%%%%%%%%%%%%%%%%%%%%%%%%%%%%%%%%%%%%%%%

\bibliographystyle{IEEEtranBST/IEEEtran_new}
\bibliography{IEEEabrv,root}

\end{document}